\def\year{2022}\relax
\documentclass[letterpaper]{article} 
\usepackage{aaai22}  
\usepackage{times}  
\usepackage{helvet}  
\usepackage{courier}  
\usepackage[hyphens]{url}  
\usepackage{graphicx} 
\usepackage{natbib}  
\usepackage{caption} 
\DeclareCaptionStyle{ruled}{labelfont=normalfont,labelsep=colon,strut=off} 
\usepackage{algorithm}
\usepackage{algorithmic}

\usepackage{newfloat}
\usepackage{listings}
\floatstyle{ruled}
\newfloat{listing}{tb}{lst}{}
\floatname{listing}{Listing}
\usepackage{latexsym}

\usepackage[utf8]{inputenc}

\usepackage{microtype}

\usepackage{xcolor}
\usepackage{booktabs}
\definecolor{light-gray}{gray}{0.95}
\definecolor{codegreen}{rgb}{0,0.6,0}
\definecolor{codegray}{rgb}{0.5,0.5,0.5}
\definecolor{codepurple}{rgb}{0.58,0,0.82}
\definecolor{backcolour}{rgb}{0.95,0.95,0.92}
\definecolor{deepblue}{rgb}{0,0,0.5}
\definecolor{deepred}{rgb}{0.6,0,0}
\definecolor{deepgreen}{rgb}{0,0.5,0}

\lstdefinestyle{mystyle}{
    language=Python,
    backgroundcolor=\color{backcolour}, commentstyle=\color{deepgreen},
    keywordstyle=\bfseries\color{codegreen},
    numberstyle=\tiny\color{codegray},
    stringstyle=\color{codepurple},
    basicstyle=\ttfamily\footnotesize,
    emphstyle=\bfseries\color{deepblue},    
    emph={boostsa, True, False},
    breakatwhitespace=false,         
    breaklines=true,                 
    captionpos=b,                    
    keepspaces=true,                 
    numbers=left,                    
    numbersep=5pt,                  
    showspaces=false,                
    showstringspaces=false,
    showtabs=false,                  
    tabsize=2
}

\title{Leveraging Social Interactions to Detect Misinformation on Social Media}
\author{Tommaso Fornaciari\textsuperscript{\rm 1}, Luca Luceri\textsuperscript{\rm 2}, Emilio Ferrara\textsuperscript{\rm 3}, Dirk Hovy\textsuperscript{\rm 4}\\
  }
\affiliations{
\textsuperscript{\rm 1}Italian National Police\\
\textsuperscript{\rm 2, \rm 3}University of Southern California\\
\textsuperscript{\rm 4}Bocconi University\\

\textsuperscript{\rm 1}\footnotesize{\texttt{tommaso.fornaciari@poliziadistato.it}}, \hspace{2mm}\textsuperscript{\rm 2}\footnotesize{\texttt{lluceri@isi.edu}},\\
\hspace{2mm}\textsuperscript{\rm 3}\footnotesize{\texttt{emiliofe@usc.edu}}, \hspace{2mm}\textsuperscript{\rm 4}\footnotesize{\texttt{dirk.hovy@unibocconi.it}}
}

\usepackage{bibentry}

\begin{document}

\maketitle

\begin{abstract}
Detecting misinformation threads is crucial to guarantee a healthy environment on social media.
We address the problem using the data set created
during the COVID-19 pandemic.
It contains cascades of tweets discussing information weakly labeled as \emph{reliable} or  \emph{unreliable}, based on a previous evaluation of the information source.
The models identifying unreliable threads usually rely on textual features.
But reliability is not just \textit{what} is said, but by whom and to whom. 
We additionally leverage on network information.
Following the homophily principle, we hypothesize that users who interact are generally interested in similar topics and spreading similar kind of news, which in turn is generally reliable or not.
We test several methods to learn representations of the social interactions within the cascades, combining them with deep neural language models in a Multi-Input (MI) framework.
Keeping track of the sequence of the interactions during the time, 
we improve over previous state-of-the-art models.
\end{abstract}

\section{Introduction}
\label{sec:intro}
Social media networks allow the wide and fast diffusion of pieces of information, news, and opinions among interacting users. However, during the last decade, the veracity and accuracy of the shared content have been largely undermined by various factors, including fake accounts and orchestrated disinformation campaigns. Fact-checking the reliability of the shared messages represents nowadays a fundamental need to preserve the integrity of online discussions and healthy fruition of social media services.

Automatically detecting misinformation spreading on social media is, however, a challenging task, as proved by the research community \cite{sharma2019combating}. Existing solutions show promising results in the classification of \emph{reliable} and \emph{unreliable} content leveraging 
the text of the shared messages.
Here we identify the threads on Twitter according to the notion of \emph{cascade}, as defined by \citet{yang2010modeling}: a sequence of reciprocally engaged tweets, ordered by their time-stamp, starting from a source post at the origin of the sequence.
We denote the cascades' reliability and unreliability according to the data set of \citet{sharma2022construction}, where the decision is mainly taken relying on an \emph{a priori} reliability evaluation of the source that issued the first tweet of the cascade.

\paragraph{Contributions}
In this paper we combine pretrained-Language Models and network-based methods, in previous literature applied to other tasks, to identify unreliable tweet cascades.
We reach new SOTA performance levels for this task.
We show how unreliable news are 1) generally associated with different communities, which can be identified and leveraged for inference,
and 2) blended with reliable news in content and style, which might not necessarily carry a strong signal.

\section{Related work}
\label{sec:relwork}
In the last ten years, several methods have been applied to the identification of unreliable cascades.
\citet{kumar2014detecting} followed a cognitive psychology approach.
\citet{zhang:detecting} work on time constraints to identify them.
\citet{yu2017convolutional} rely on textual data and propose the use of a convolutional neural network.
Also \citet{monti2019fake} rely on convolutional neural networks, but they are interested in propagation patterns, that is the geometry of the social networks that share news. A similar, hierarchical approach is followed by \citet{shu2020hierarchical}.

Deep Learning methods, such as LSTM, are applied to the texts by \citet{ducci:cascade}, \citet{pierri2020multi}.
This approach is similar to that applied by \citet{sharma2022construction}, who used the CSI model of \citet{natali:csi}, which employs a recurrent neural network to represent texts and user behaviors.
To capture social interactions, we use mentions2vec - M2V, a method proposed by \citet{fornaciari-hovy-2019-dense} and applied on a geolocation task.


\section{Data}
The data set of \citet{sharma2022construction}, collected during the COVID-19 pandemic, contains 14\,644 cascades (10\,377 reliable, 4\,267 unreliable), already divided in training, development and test set. 
The cascades contain 376\,228 tweets, issued by 168\,227 users.
The texts are in English and already pre-processed.

\section{Methods}
\label{sec:meth}
We implement five different models to detect misinformation tweet cascades.
The first is the baseline, to which we compare the four other models.
All models perform the same classification task.

The baseline model is a text-only Single-Input BERT-based \cite{devlin2018bert} model.
It uses the contextual word embeddings from BERT, without fine-tuning the whole BERT.
In particular, we use the mean of the word vectors of the concatenated tweets from the whole cascade.
However, between BERT's output and the standard, fully-connected classification layer, we insert a further Transformer mechanism \cite{vaswani2017attention}.
This approach has proven more effective than using a fully-connected output layer alone, in several NLP tasks
\cite{fornaciari-etal-2021-milanlp}.  

Similarly to \citet{sharma2022construction}, who use the same kind of inputs, we explore four different types of Multi-Input models, fed with different combinations of textual and network-interaction information.
The textual data are represented via the BERT-based language model, like in the baseline model.
The network interactions are encoded via three different methods, as follows.

\paragraph{Multi-Input: network-sparse-vectors}
The simplest way to represent a cascade in a social network as a vector is to encode all users' presence or absence in each tweet cascade.
To keep the vectors within a manageable size and reduce the noise from uninformative data (e.g., cascades with few or infrequent users), we only considered users that performed at least 15 actions (i.e., tweets, retweets, replies, or quotes) in one or more cascades in the dataset.

We chose the threshold of 15 based on computational affordability (see last paragraph in this Section).
This method produces sparse vectors of size 1326 for each cascade. The dimensions correspond to the 1326 selected users, with the values 1 if the user is present in the cascade, and 0 otherwise.
In this model, both the textual (BERT) and network (sparse) representations are separately fed into two Transformers \cite{vaswani2017attention}, whose outputs are concatenated and passed to the final classification layer.

\paragraph{Multi-Input: network-embeddings}
In the second model, the sparse vectors conveying the network interactions view are not passed directly to an attention mechanism but are fed into a fully connected layer, which squeezes them into dense vectors of size 128.
These smaller, dense vectors can be considered learned (i.e., trainable) network embeddings.
Then, similarly to the previous models, network and text (BERT) embeddings are passed to two parallel attention mechanisms connected to the classification layer.

\begin{figure*}
    \includegraphics[width=\textwidth]{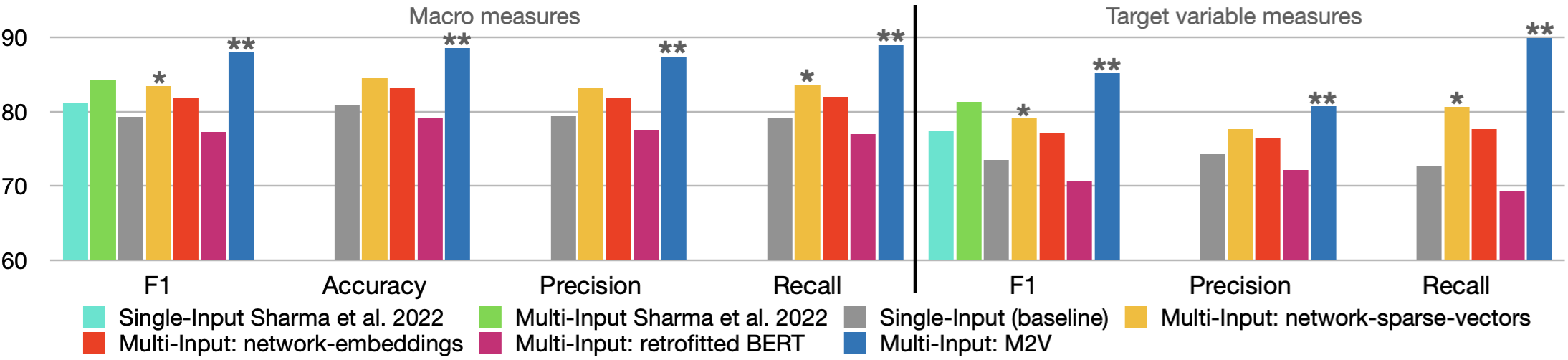}
    \caption{Overall and target class (i.e., unreliable cascades) performance. Significance: $^{\ast\ast}: p \le 0.01; \hspace{2mm} ^{\ast}: p \le 0.05$.} 
    \label{fig:res}
\end{figure*} 

\paragraph{Multi-Input: mentions2vec - M2V network-embeddings}
In the third model, we again use the textual BERT representations as in the previous models.
For the network representation, we use mentions2vec, a methods based on Doc2Vec \cite{d2v} proposed by \citet{fornaciari-hovy-2019-dense} (there to improve model performance in a geolocation task).
M2V filters the texts to preserve only the users' mentions (i.e., user names starting with ``@'').
This procedure results in ``texts'' containing only sequences of users' mentions. In this way, the texts represent explicit social interactions on social media.
These sequences are then encoded as dense vectors using Doc2Vec. Doc2Vec allows for the assignment of document labels, typically the document ID. Here, we substituted this label with the cascade ID.

This procedure has a critical advantage over the other, traditional methods of network representation. Those rely on square (adjacency) matrices that grow quadratically with the network size, a constraint that quickly becomes computationally unsustainable. 
Therefore, other methods need to keep rigid control of the network size and typically revert to some form of sampling when the network size becomes too large.
M2V, in contrast, produces fixed-length vectors of a chosen size, independent of network size. The number of users does not affect the size of the representations, and the number of user mentions acts as a ``vocabulary'' in Doc2Vec. 

This way the growth of the network representation is linear with the number of texts, rather than quadratic with the number of users.
We feed the M2V network embeddings into the same architecture used for the previous experimental models. 

\paragraph{Multi-Input: retrofitted-BERT and network-embeddings}
In the last model, we use the same network-embedding representations as before.
However, we evaluate the possibility of injecting cascade information into the BERT embeddings. 
We do this via the cascade classes in the training set, and use them in the retrofitting method proposed by \citet{faruqui:retro}.
It forces the vectors of instances belonging to the same equivalence class (here, the cascade class) to be more similar to each other, thereby increasing the distance between instances of different classes.
This kind of transformation can be reproduced on unseen data, even if the class is unknown, using a translation matrix that approximates the original operation of matrix transformation \cite{faruqui:retro,hovy-fornaciari-2018-increasing}.

In our case, we retrofit the texts' representations of the training data according to their relative cascade label. I.e., we start with BERT mean word embeddings, and increase the similarity of all vectors that represent reliable cases, and the similarity of all vectors labeled unreliable.
Then we learn a translation matrix from the original BERT embeddings to the retrofitted ones, and apply the translation to the development and test data to  create a retrofitted version of them. This second step does no longer require access to labels: the translation matrix has learned how to transform textual embeddings to reflect cascade classes.

Retrofitting the data is a form of pre-processing, as it precedes training and is not affected by the model training.
We use the same neural architecture of the previous model, but fed with retrofitted rather than standard textual embeddings.
Strictly speaking, this procedure does not leverage network information directly, as it relies on cascade labels.
However, this approach lets us verify whether we can improve model performance by leveraging the association between texts based on their label.

\paragraph{Computational load, parameters and hyper-parameters.}
In our experiments, we used a GPU NVIDIA GeForce GTX 1080 Ti.
The BERT models were not fine-tuned. 
The number of trainable parameters ranged from 8.5M for the Single-Input models to 20M for the mentions2vec-based models.
To reduce the random initializations' impact, we carried out five experiments for each experimental condition.
The creation of the BERT text representation took 40 minutes, and the set of the following experiments was approximately one hour.
The training was stopped with an early stopping algorithm, relying on the development set's F-measure. 
The tables in Appendix show the mean epochs for each experimental condition.
We used Transformers with one layer and one head, dropout probability for the classification layer at .1. These hyper-parameters were found through empirical search.

\section{Results}
\label{sec:res}
Figure \ref{fig:res} show the results.
The left side shows the macro performance, that is the overall performance averaged over the two classes.
The right side focuses on the target class, that is the performance on predicting unreliable cascades.
In both cases, we compare our results to the performance of the best previously-reported models \cite{sharma2022construction}.
Following common good practice in NLP, we use bootstrap sampling \cite{efron1994bootstrap,berg2012significance} to compute the performance significance between the Multi-Input models and the Single-Input baseline.
We repeat 1000 tests per model, with a sampling size of 30\% \cite{sogaard,fornaciari-etal-2022-hard}.

The models of \citet{sharma2022construction} are challenging to beat.
Their Single-Input model handily beats our corresponding baseline model. Their Multi-Input model tends to be better than most formulations we explore.
However, our Multi-Input model using M2V manages to significantly improve over the baseline with $p \le 0.01$, and it improves by more than 3.5 points F1 over the best model of \citet{sharma2022construction} in both settings (the macro value and the target class only, see Appendix).  

By comparing our Multi-Input models against the Single-Input ones, which all share the same textual representation, we can measure the specific network representation's contribution to the classification task.
The models relying on network-sparse-vectors are significantly better than 
our Single-Input baseline.
The models with network embeddings are still better than the control, but not significantly.

Lastly, the models that use retrofitted BERT embeddings show results that are even worse than the baseline, which did not incorporate network information.

\section{Discussion}
\label{sec:disc}
The low performance of the model with retrofitted BERT embeddings is an interesting result. Making the cascade representations from the same label class more similar does not improve performance.
This outcome suggests that, from the point of view of style and content, reliable and unreliable cascades are quite similar to each other.
\emph{Ex post}, it makes sense that topics completely different from each other could still share the same feature of being reliable or unreliable.
In contrast, network representations are clearly useful for classification.
We assume that different communities are prone to congregate around different topics that tend to be systematically more reliable or unreliable.
Feeding sparse cascade vectors directly into the Transformers is a strategy more effective than previously reducing their dimension with a dense representation.

Since the dense representation approximates the sparse one, the results are not surprising.
However, for wider networks, feeding sparse vectors into Transformers that rely on multiple `key', `query', and `value' square matrices could be computationally unaffordable \cite{vaswani2017attention}.

Finally, the M2V approach proves particularly effective for the task. 
The information modeled in this representation is much richer than that simply inferred by counting the users present in the same cascade. 
M2V considers all the accounts a user addresses in the texts that he/she produces.
This set of accounts can also include `silent' users and so can be (much) wider than the group of users who actively participate in the cascade.
This means that the social representation is particularly expressive.
Also, users can be mentioned several times, which would give their presence (or influence) more weight in the representation.   

\begin{figure}
    \includegraphics[width=\linewidth]{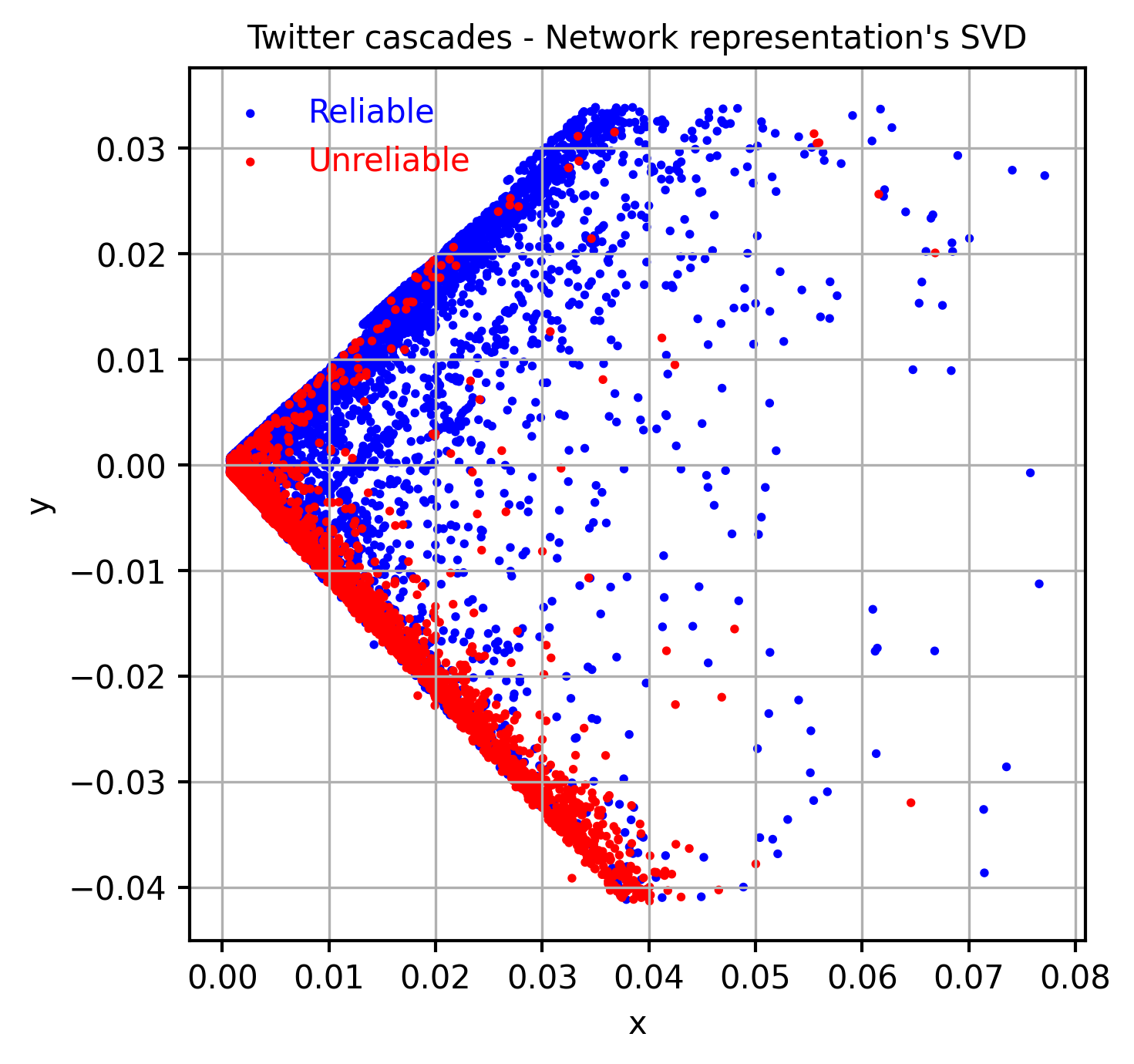}
    \caption{Twitter cascades in the test data represented by their users via SVD. Color shows label class.} 
    \label{fig:svd}
\end{figure}

\section{User clustering}
\label{sec:clus}
To test our hypothesis that reliable and unreliable cascades are really fed by different communities, we applied unsupervised methods to cluster the cascades according to the users who wrote texts in the cascades.
In particular, we vectorized the cascades via the method used to create the sparse network representation, but without filtering the users according to the frequency threshold of 15, used in that case (Section Methods, paragraph Multi-Input: network-sparse-vectors).
Then, we reduced the vectors' dimensionality with Truncated Singular Value Decomposition - SVD \cite[Chapter 18]{sanderson2010christopher}.

The results, without outliers (which would have ``zoomed out'' the whole image), are shown in Figure \ref{fig:svd}.
Reliable and unreliable cascades are clearly positioned in different regions of the chart, suggesting that they are characterized by the presence of different users. This finding points towards a community-driven aspect of reliability.

\section{Conclusion}
\label{sec:conc}
In this paper, we explored four computational methods to detect unreliable tweet cascades.
Our results suggest that these harmful threads can contain various topics; however, they mostly are generated by distinct communities.

Therefore it is useful to support the linguistic representations with a network view, which proves effective for this task.
Among other methods, we find that mentions2vec \cite{fornaciari-hovy-2019-dense} is an efficient way to encode user interactions within the cascades.

As recent research demonstrates that some users play a pivotal role in diffusing questionable information~\cite{nogara2022disinformation, yang2021covid, deverna2022superspreaders}, in future work, we will develop solutions to embed in our models the activity of the so-called misinformation ``superspreaders''.

\section*{Limitations}
English is the target language of this study. Reproducibility might be problematic with languages with wider morphology.
Also, the presented methods of social network analysis require data from social media that allow mentioning unambiguously other user accounts with some markup sign, ``@'' in the Twitter case. 

\section*{Ethical Considerations}
We adopted publicly available datasets for training and testing our framework but we did not devote sufficient time and attention to the possible biases that our model might have that could yield practical implications in real-world applications. 
We do not believe our framework is harmful \emph{per se}.
However, as the input documents and their representations might carry biases, unethical content, and/or personal information, issues of fairness, bias, and data privacy might arise.
Therefore, we invite further research and responsible use of this framework.


\section*{Acknowledgments}
Work supported in part by DARPA (contract \#HR001121C0169).

\bibliography{custom}

\begin{thebibliography}{26}
\providecommand{\natexlab}[1]{#1}

\bibitem[{Berg-Kirkpatrick, Burkett, and Klein(2012)}]{berg2012significance}
Berg-Kirkpatrick, T.; Burkett, D.; and Klein, D. 2012.
\newblock An Empirical Investigation of Statistical Significance in {NLP}.
\newblock In \emph{Proceedings of the 2012 Joint Conference on Empirical
  Methods in Natural Language Processing and Computational Natural Language
  Learning}, 995--1005. Jeju Island, Korea: Association for Computational
  Linguistics.

\bibitem[{DeVerna et~al.(2022)DeVerna, Aiyappa, Pacheco, Bryden, and
  Menczer}]{deverna2022superspreaders}
DeVerna, M.~R.; Aiyappa, R.; Pacheco, D.; Bryden, J.; and Menczer, F. 2022.
\newblock Identification and characterization of misinformation superspreaders
  on social media.
\newblock \emph{arXiv preprint arXiv:2207.09524}.

\bibitem[{Devlin et~al.(2018)Devlin, Chang, Lee, and
  Toutanova}]{devlin2018bert}
Devlin, J.; Chang, M.-W.; Lee, K.; and Toutanova, K. 2018.
\newblock Bert: Pre-training of deep bidirectional transformers for language
  understanding.
\newblock \emph{arXiv preprint arXiv:1810.04805}.

\bibitem[{Ducci, Kraus, and Feuerriegel(2020)}]{ducci:cascade}
Ducci, F.; Kraus, M.; and Feuerriegel, S. 2020.
\newblock Cascade-LSTM: A Tree-Structured Neural Classifier for Detecting
  Misinformation Cascades.
\newblock In \emph{Proceedings of the 26th ACM SIGKDD International Conference
  on Knowledge Discovery; Data Mining}, KDD '20, 2666–2676. New York, NY,
  USA: Association for Computing Machinery.
\newblock ISBN 9781450379984.

\bibitem[{Efron and Tibshirani(1994)}]{efron1994bootstrap}
Efron, B.; and Tibshirani, R.~J. 1994.
\newblock \emph{An introduction to the bootstrap}.
\newblock CRC press.

\bibitem[{Faruqui et~al.(2015)Faruqui, Dodge, Jauhar, Dyer, Hovy, and
  Smith}]{faruqui:retro}
Faruqui, M.; Dodge, J.; Jauhar, S.~K.; Dyer, C.; Hovy, E.; and Smith, N.~A.
  2015.
\newblock Retrofitting Word Vectors to Semantic Lexicons.
\newblock In \emph{Proceedings of the 2015 Conference of the North American
  Chapter of the Association for Computational Linguistics: Human Language
  Technologies}, 1606--1615.

\bibitem[{Fornaciari et~al.(2021)Fornaciari, Bianchi, Nozza, and
  Hovy}]{fornaciari-etal-2021-milanlp}
Fornaciari, T.; Bianchi, F.; Nozza, D.; and Hovy, D. 2021.
\newblock {M}ila{NLP} @ {WASSA}: Does {BERT} Feel Sad When You Cry?
\newblock In \emph{Proceedings of the Eleventh Workshop on Computational
  Approaches to Subjectivity, Sentiment and Social Media Analysis}, 269--273.
  Online: Association for Computational Linguistics.

\bibitem[{Fornaciari and Hovy(2019)}]{fornaciari-hovy-2019-dense}
Fornaciari, T.; and Hovy, D. 2019.
\newblock Dense Node Representation for Geolocation.
\newblock In \emph{Proceedings of the 5th Workshop on Noisy User-generated Text
  (W-NUT 2019)}, 224--230. Hong Kong, China: Association for Computational
  Linguistics.

\bibitem[{Fornaciari et~al.(2022)Fornaciari, Uma, Poesio, and
  Hovy}]{fornaciari-etal-2022-hard}
Fornaciari, T.; Uma, A.; Poesio, M.; and Hovy, D. 2022.
\newblock Hard and Soft Evaluation of {NLP} models with {BOO}t{ST}rap
  {SA}mpling - {B}oo{S}t{S}a.
\newblock In \emph{Proceedings of the 60th Annual Meeting of the Association
  for Computational Linguistics: System Demonstrations}, 127--134. Dublin,
  Ireland: Association for Computational Linguistics.

\bibitem[{Hovy and Fornaciari(2018)}]{hovy-fornaciari-2018-increasing}
Hovy, D.; and Fornaciari, T. 2018.
\newblock Increasing In-Class Similarity by Retrofitting Embeddings with
  Demographic Information.
\newblock In \emph{Proceedings of the 2018 Conference on Empirical Methods in
  Natural Language Processing}, 671--677. Brussels, Belgium: Association for
  Computational Linguistics.

\bibitem[{Kumar and Geethakumari(2014)}]{kumar2014detecting}
Kumar, K.; and Geethakumari, G. 2014.
\newblock Detecting misinformation in online social networks using cognitive
  psychology.
\newblock \emph{Human-centric Computing and Information Sciences}, 4(1): 1--22.

\bibitem[{Le and Mikolov(2014)}]{d2v}
Le, Q.; and Mikolov, T. 2014.
\newblock Distributed representations of sentences and documents.
\newblock In \emph{International Conference on Machine Learning}, 1188--1196.

\bibitem[{Monti et~al.(2019)Monti, Frasca, Eynard, Mannion, and
  Bronstein}]{monti2019fake}
Monti, F.; Frasca, F.; Eynard, D.; Mannion, D.; and Bronstein, M.~M. 2019.
\newblock Fake news detection on social media using geometric deep learning.
\newblock \emph{arXiv preprint arXiv:1902.06673}.

\bibitem[{Nogara et~al.(2022)Nogara, Vishnuprasad, Cardoso, Ayoub, Giordano,
  and Luceri}]{nogara2022disinformation}
Nogara, G.; Vishnuprasad, P.~S.; Cardoso, F.; Ayoub, O.; Giordano, S.; and
  Luceri, L. 2022.
\newblock The Disinformation Dozen: An Exploratory Analysis of Covid-19
  Disinformation Proliferation on Twitter.
\newblock In \emph{14th ACM Web Science Conference 2022}, 348--358.

\bibitem[{Pierri, Piccardi, and Ceri(2020)}]{pierri2020multi}
Pierri, F.; Piccardi, C.; and Ceri, S. 2020.
\newblock A multi-layer approach to disinformation detection in US and Italian
  news spreading on Twitter.
\newblock \emph{EPJ Data Science}, 9(1): 35.

\bibitem[{Ruchansky, Seo, and Liu(2017)}]{natali:csi}
Ruchansky, N.; Seo, S.; and Liu, Y. 2017.
\newblock CSI: A Hybrid Deep Model for Fake News Detection.
\newblock In \emph{Proceedings of the 2017 ACM on Conference on Information and
  Knowledge Management}, CIKM '17, 797–806. New York, NY, USA: Association
  for Computing Machinery.
\newblock ISBN 9781450349185.

\bibitem[{Sanderson(2010)}]{sanderson2010christopher}
Sanderson, M. 2010.
\newblock Christopher D. Manning, Prabhakar Raghavan, Hinrich Sch{\"u}tze,
  Introduction to Information Retrieval, Cambridge University Press. 2008.
  ISBN-13 978-0-521-86571-5, xxi+ 482 pages.
\newblock \emph{Natural Language Engineering}, 16(1): 100--103.

\bibitem[{Sharma, Ferrara, and Liu(2022)}]{sharma2022construction}
Sharma, K.; Ferrara, E.; and Liu, Y. 2022.
\newblock Construction of Large-Scale Misinformation Labeled Datasets from
  Social Media Discourse using Label Refinement.
\newblock In \emph{Proceedings of the ACM Web Conference 2022}, 3755--3764.

\bibitem[{Sharma et~al.(2019)Sharma, Qian, Jiang, Ruchansky, Zhang, and
  Liu}]{sharma2019combating}
Sharma, K.; Qian, F.; Jiang, H.; Ruchansky, N.; Zhang, M.; and Liu, Y. 2019.
\newblock Combating fake news: A survey on identification and mitigation
  techniques.
\newblock \emph{ACM Transactions on Intelligent Systems and Technology (TIST)},
  10(3): 1--42.

\bibitem[{Shu et~al.(2020)Shu, Mahudeswaran, Wang, and
  Liu}]{shu2020hierarchical}
Shu, K.; Mahudeswaran, D.; Wang, S.; and Liu, H. 2020.
\newblock Hierarchical propagation networks for fake news detection:
  Investigation and exploitation.
\newblock In \emph{Proceedings of the international AAAI conference on web and
  social media}, volume~14, 626--637.

\bibitem[{S{\o}gaard et~al.(2014)S{\o}gaard, Johannsen, Plank, Hovy, and
  Mart{\'\i}nez~Alonso}]{sogaard}
S{\o}gaard, A.; Johannsen, A.; Plank, B.; Hovy, D.; and Mart{\'\i}nez~Alonso,
  H. 2014.
\newblock What{'}s in a p-value in {NLP}?
\newblock In \emph{Proceedings of the Eighteenth Conference on Computational
  Natural Language Learning}, 1--10. Ann Arbor, Michigan: Association for
  Computational Linguistics.

\bibitem[{Vaswani et~al.(2017)Vaswani, Shazeer, Parmar, Uszkoreit, Jones,
  Gomez, Kaiser, and Polosukhin}]{vaswani2017attention}
Vaswani, A.; Shazeer, N.; Parmar, N.; Uszkoreit, J.; Jones, L.; Gomez, A.~N.;
  Kaiser, {\L}.; and Polosukhin, I. 2017.
\newblock Attention is all you need.
\newblock \emph{Advances in neural information processing systems}, 30.

\bibitem[{Yang and Leskovec(2010)}]{yang2010modeling}
Yang, J.; and Leskovec, J. 2010.
\newblock Modeling information diffusion in implicit networks.
\newblock In \emph{2010 IEEE International Conference on Data Mining},
  599--608. IEEE.

\bibitem[{Yang et~al.(2021)Yang, Pierri, Hui, Axelrod, Torres-Lugo, Bryden, and
  Menczer}]{yang2021covid}
Yang, K.-C.; Pierri, F.; Hui, P.-M.; Axelrod, D.; Torres-Lugo, C.; Bryden, J.;
  and Menczer, F. 2021.
\newblock The COVID-19 Infodemic: Twitter versus Facebook.
\newblock \emph{Big Data \& Society}, 8(1): 20539517211013861.

\bibitem[{Yu et~al.(2017)Yu, Liu, Wu, Wang, Tan et~al.}]{yu2017convolutional}
Yu, F.; Liu, Q.; Wu, S.; Wang, L.; Tan, T.; et~al. 2017.
\newblock A Convolutional Approach for Misinformation Identification.
\newblock In \emph{IJCAI}, 3901--3907.

\bibitem[{Zhang et~al.(2016)Zhang, Kuhnle, Zhang, and Thai}]{zhang:detecting}
Zhang, H.; Kuhnle, A.; Zhang, H.; and Thai, M.~T. 2016.
\newblock Detecting misinformation in online social networks before it is too
  late.
\newblock In \emph{2016 IEEE/ACM International Conference on Advances in Social
  Networks Analysis and Mining (ASONAM)}, 541--548.

\end{thebibliography}

\appendix

\section{Appendix}

\begin{table*}
\centering
\small
\begin{tabular}{lllllll}
\toprule
Model & Mean & Macro-F1 & Std. & Accuracy & Precision & Recall \\
 & epochs & & Dev. & & & \\
\midrule
Single-Input \citet{sharma2022construction}: Weak labels & & 81.20 & 0.01 & & & \\
Multi-Input \citet{sharma2022construction}: Social+Detection model & & 84.20 & 0.01 & & & \\
\midrule
Single-Input (baseline) & 14.20 & 79.30  & 0.01 & 80.94  & 79.45  & 79.17  \\
Multi-Input: network-sparse-vectors & 13.20 & 83.42 * & 0.02 & 84.53  & 83.17  & 83.70 * \\
Multi-Input: network-embeddings & 23.60 & 81.90  & 0.01 & 83.20  & 81.81  & 82.01  \\
Multi-Input: retrofitted BERT + network-embeddings & 11.40 & 77.25  & 0.01 & 79.14  & 77.53  & 77.02  \\
Multi-Input: M2V & 10.00 & \textbf{87.95 **} & 0.00 & \textbf{88.59 **} & \textbf{87.34 **} & \textbf{88.92 **} \\
\bottomrule
\end{tabular}
\caption{Overall performance on detection task. Significance: $^{\ast\ast}: p \le 0.01; \hspace{2mm} ^{\ast}: p \le 0.05$. \textbf{Bold}: best column result.}
\label{tab:macro}
\end{table*}

\begin{table*}
\centering
\small
\begin{tabular}{llllll}
\toprule
Model & Mean & F1 & Std. & Precision & Recall \\
 & epochs & & Dev. & & \\
\midrule
Single-Input \citet{sharma2022construction}: Weak labels & & 77.40 & 0.02 & & \\
Multi-Input \citet{sharma2022construction}: Social+Detection model & & 81.30 & 0.01 & & \\
\midrule
Single-Input (baseline) & 14.20 & 73.48  & 0.02 & 74.29  & 72.69  \\
Multi-Input: network-sparse-vectors & 13.20 & 79.11 * & 0.03 & 77.64  & 80.65 * \\
Multi-Input: network-embeddings & 23.60 & 77.05  & 0.02 & 76.48  & 77.63  \\
Multi-Input: retrofitted BERT + network-embeddings & 11.40 & 70.69  & 0.02 & 72.20  & 69.25  \\
Multi-Input: M2V & 10.00 & \textbf{85.16 **} & 0.00 & \textbf{80.73 *} & \textbf{90.11 **} \\
\bottomrule
\end{tabular}
\caption{Performance on detecting target class \textit{only} (i.e., unreliable cascades). Significance: $^{\ast\ast}: p \le 0.01; \hspace{2mm} ^{\ast}: p \le 0.05$. \textbf{Bold}: best column result.}
\label{tab:target}
\end{table*}


\end{document}